\pdfoutput=1

\documentclass[11pt]{article}

\usepackage{emnlp2021}

\usepackage{times}
\usepackage{latexsym}

\usepackage[T1]{fontenc}

\usepackage[utf8]{inputenc}

\usepackage{microtype}

\usepackage{multirow}
\usepackage{subfigure}
\usepackage{bm}
\usepackage{amssymb}
\usepackage{verbatim}
\usepackage{graphicx}
\usepackage{amsmath}
\usepackage{amsthm}
\usepackage{booktabs}
\usepackage[noend]{algpseudocode}
\usepackage{algorithmicx,algorithm}
\usepackage{stfloats}
\title{Efficient Mind-Map Generation via Sequence-to-Graph and Reinforced \\ Graph Refinement}

\author{Mengting Hu\textsuperscript{1,2} \quad Honglei Guo\textsuperscript{3}\thanks{\; Honglei Guo is the corresponding author.} \quad Shiwan Zhao\textsuperscript{3} \quad {\bf Hang Gao\textsuperscript{4}  \quad Zhong Su\textsuperscript{3}} \\
\textsuperscript{1} College of Software, Nankai University \quad
\textsuperscript{2} Tianjin Key Laboratory of Operating System \\
\textsuperscript{3} IBM Research - China \quad
\textsuperscript{4} Institute for Public Safety Research, Tsinghua University \\
{\tt mthu@nankai.edu.cn,} {\tt gaohang@mail.tsinghua.edu.cn} \\
{\tt \{guohl,zhaosw,suzhong\}@cn.ibm.com} 
}

\begin{document}
\maketitle
\begin{abstract}
A mind-map is a diagram that represents the central concept and key ideas in a hierarchical way. Converting plain text into a mind-map will reveal its key semantic structure and be easier to understand. Given a document, the existing automatic mind-map generation method extracts the relationships of every sentence pair to generate the directed semantic graph for this document. The computation complexity increases exponentially with the length of the document. Moreover, it is difficult to capture the overall semantics. To deal with the above challenges, we propose an efficient mind-map generation network that converts a document into a graph via sequence-to-graph. To guarantee a meaningful mind-map, we design a graph refinement module to adjust the relation graph in a reinforcement learning manner. Extensive experimental results demonstrate that the proposed approach is more effective and efficient than the existing methods. The inference time is reduced by thousands of times compared with the existing methods. The case studies verify that the generated mind-maps better reveal the underlying semantic structures of the document.
\end{abstract}



\section{Introduction}
A mind-map is a hierarchical diagram that can depict the central concept, linked with the major ideas and other ideas branch out from these \cite{kudelic2011mind,ijcai2019-729}. It is organized in cognitive structures and much easier to understand than plain text \cite{dhindsa2011constructivist}. Thus in practice, it can be utilized for education resources, organizing, and planning. Many tools can help people \emph{make} mind-map manually, such as FreeMind, MindGenius and Visual Mind, etc \cite{kudelic2011mind}. To save human labors, some automatic methods have been proposed to \emph{generate} mind-map from text, which focus on analyzing the semantic relations \emph{within a sentence} by pre-defined rules \cite{brucks2008assembling,rothenberger2008figuring} or syntactic parser \cite{elhoseiny2012english2mindmap}. 

\begin{figure}[t]
\centering
\includegraphics[width=0.48\textwidth]{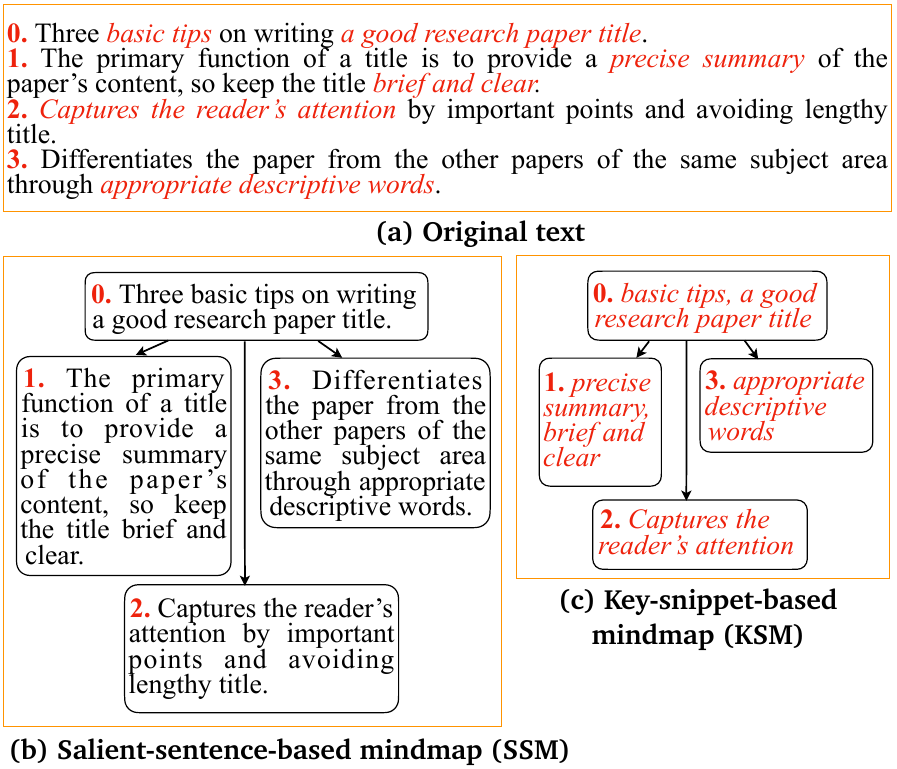}
\caption{The original text (a) is converted into mind-maps, in which a node is the entire sentence (b) or keywords (c).}
\label{example}
\end{figure}

Recently, researchers \cite{ijcai2019-729} propose to generate a mind-map automatically by detecting the semantic relation \emph{cross sentences} in the document. It mines the structured diagram of the document, in which a node represents the meaning of a sentence in the format of the entire sentence or its keywords, and an edge represents the governing relationships between the precursor and the successor. We illustrate two types of mind-map in Figure \ref{example}, i.e. salient-sentence-based mind-map (SSM) and key-snippet-based mind-map (KSM).

\begin{figure}[t]
\centering
\includegraphics[width=0.48\textwidth]{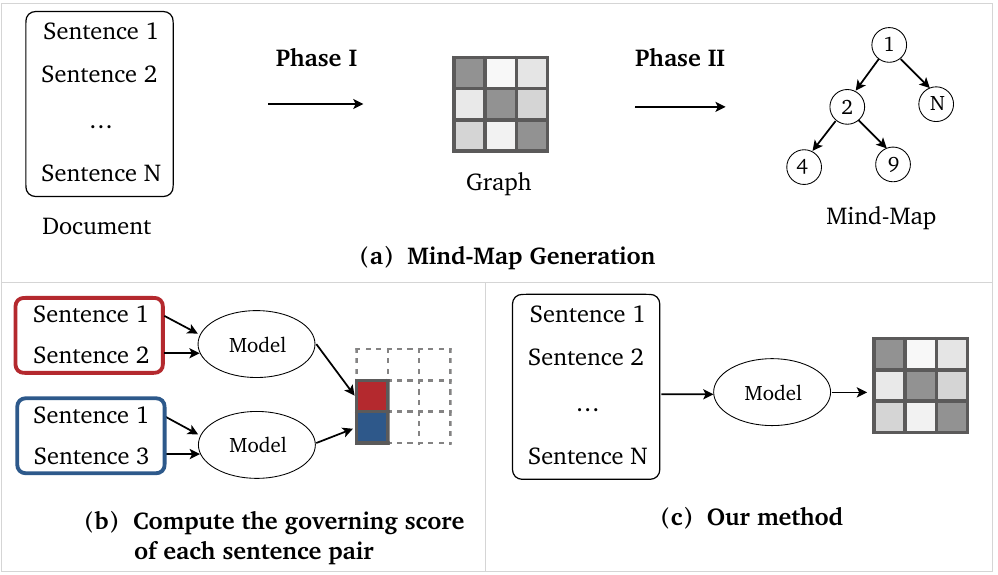}
\caption{(a) Mind-map generation procedure. (b) MRDMF \protect\cite{ijcai2019-729} predicts the relation graph at the sentence pair level. (c) Our method predicts the relation graph at the document level.}
\label{example_method}
\end{figure}

\newcite{ijcai2019-729} propose a promising pipeline approach (see Figure \ref{example_method}(a)), which first converts the whole document into a relation graph and then prunes the extra edges to obtain the mind-map. However, the first phase tends to be less efficient since it needs to predict all the governing scores at the sentence pair level (see Figure \ref{example_method}(b)). The number of sentence pairs increases exponentially with the length of the document, which raises the computational complexity. In addition, each governing score in the graph is computed separately without considering the overall semantics in the document. The sequential information of all sentences might be helpful to mine the hierarchical and structured semantics.

We propose an efficient mind-map generation network (EMGN) to address the above issues (see Figure \ref{example_method}(c)). The proposed method encodes all sentences sequentially and generates the graph via sequence-to-graph. It makes the first phase more efficient and can easily process multiple documents in parallel. The model training requires all the relation labels of sentence pairs in each graph. However, manually annotating costs much. We exploit DistilBert \cite{sanh2019distilbert} to automatically annotate a graph for each document, which provides pseudo labels to train our model. In advance, DistilBert has been fine-tuned to detect the governing relation between two sentences. The performance of DistilBert indicates it can be an ``annotator'' with high confidence. 



Moreover, a meaningful mind-map tends to organize the major ideas of a document close to the root node. To achieve this goal, we design a graph refinement module to adjust the generated graph by using the documents with highlights. The highlights written by the editors summarize the key ideas of a document. During training, we leverage this prior human knowledge as a reference to refine the governing scores in the graph via self-critical reinforcement learning \cite{rennie2017self}. 



In summary, the main contributions of this paper are as follows.
\begin{itemize}
    \item We propose an efficient mind-map method that can consider the document-level semantics by sequence-to-graph.
    \item In the training phase, we design a graph refinement module to refine the generated graph by leveraging the manual highlights and self-critical reinforcement learning.
    \item Extensive experimental results demonstrate the proposed method can generate a better-performed mind-map efficiently. The inference time is reduced by thousands of times compared with the existing approaches. 
\end{itemize}

\section{Methodology}

\subsection{Problem Definition}
Assume a document has $N$ sentences $D=\{s_k\}_{k=1}^N$, each sentence is a word sequence $s_k=\{w_k^1,w_k^2,...,w_k^{L_k}\}$, where $L_k$ is the length of the sentence. We define the mind-map generation as a two-phase task.
\begin{equation}
    D\rightarrow\mathbf{G}\rightarrow{M}
\end{equation}
where the input text $D$ is first processed to obtain the relation graph $\mathbf{G}$. Then the graph is pruned to gain the final mind-map $M$. 

The detailed methodology for this two-phase task is described in the following sections. Concretely, we depict the network architecture for the first phase $D\rightarrow{\mathbf{G}}$ in Figure \ref{Network}. We generate the relation graph from a document by graph detector (\S\ref{subsec:graph_detector}). The graph is simultaneously refined to make the generated mind-map more meaningful (\S\ref{subsec:graph_refinement}). For the second phase $\mathbf{G}\rightarrow{M}$, we generate two types of mind-maps based on the graph (\S\ref{subsec:mind_map_detector}). 

\subsection{Graph Detector}
\label{subsec:graph_detector}
\begin{figure}[t]
\centering
\includegraphics[width=0.48\textwidth]{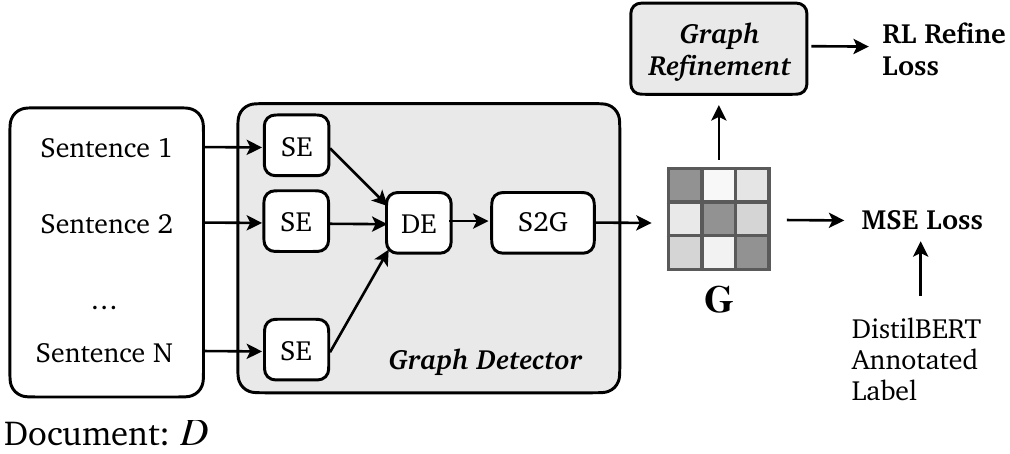}
\caption{The network architecture of the proposed approach for converting the document to a graph (Phase \uppercase\expandafter{\romannumeral1}). SE, DE, and S2G refer to the sentence encoder, document encoder, and sequence-to-graph modules, respectively.}
\label{Network}
\end{figure}

As shown in Figure \ref{Network}, the graph detector aims to extract the relation graph for an input document. It considers the overall semantics and obtains the graph efficiently. 

\subsubsection{Sentence Encoder}
Given a sentence $s_k$, we first map it into an embedding sequence $\{\bm{e_k^1,e_k^2,...,e_k^{L_k}}\}$ through a pre-trained embedding matrix GloVE \cite{pennington2014glove}. Then we exploit a Bi-directional LSTM (BiLSTM) \cite{Graves2013Speech} to encode the embedding sequence into the hidden states $\{\bm{h_k^1,h_k^2,...h_k^{L_k}}\}$. To compute the vector representation for each sentence, we apply a simple max-pooling operation over the hidden states. 
\begin{equation}
    \bm{s_k}=\mathrm{max}(\bm{h_k^1},...,\bm{h_k^{L_k}})
\end{equation}


\subsubsection{Document Encoder}
The sequential information of sentences indicates the semantic coherence and logical structure of a document. This information is essential in understanding the entire document and extracting a clear mind-map. To model the sentence-level context, we encode the vector representations of all sentences $\{\bm{s_k}\}_{k=1}^N$ with another BiLSTM and obtain $\mathbf{H}=\{\bm{h_1},\bm{h_2},...,\bm{h_N}\}$.

\subsubsection{Sequence-to-Graph}
In a graph $\mathbf{G}$, a node represents a sentence from the document. $\mathbf{G}_{i,j}$ is the governing score between sentence $s_i$ and $s_j$, which indicates the probability that $s_i$ semantically implies $s_j$. Thus the graph is directed since the governing relationships are different between $\mathbf{G}_{i,j}$ and $\mathbf{G}_{j,i}$. Inspired by \cite{dozat2016deep,zhang-etal-2019-amr}, we utilize sequence-to-graph to process the sentence-level sequence into graph efficiently. Concretely, we first compute the representations of all sentences when they are regarded as the start or end nodes in the edges. Exploiting separate parameters help learn distinct representations for a sentence.
\begin{equation}
    \begin{split}
        \bm{h_i}^{\mathrm{(start)}} &= \mathrm{MLP}^{\mathrm{(start)}}(\bm{h_i}) \\
        \bm{h_j}^{\mathrm{(end)}} &= \mathrm{MLP}^{\mathrm{(end)}}(\bm{h_j})
    \end{split}
\end{equation}
where $\mathrm{MLP}$ is a linear transformation. Then we calculate the governing scores in $\mathbf{G}\in\mathbb{R}^{N\times{N}}$ with a bilinear operation or biaffine operation. 
\begin{equation}
    \begin{split}
        \mathbf{G}_{i,j}&=\mathrm{Bilinear}(\bm{h_i}^{\mathrm{(start)}},\bm{h_j}^{\mathrm{(end)}})\\
        \mathbf{G}_{i,j}&=\mathrm{Biaffine}(\bm{h_i}^{\mathrm{(start)}},\bm{h_j}^{\mathrm{(end)}})
    \end{split}
\end{equation}
where $\mathrm{Bilinear}$ and $\mathrm{Biaffine}$ are defined as below.
\begin{equation}
\small
\begin{split}
    \mathrm{Bilinear}(\bm{x_1},\bm{x_2}) &= \mathrm{\sigma}(\bm{x_1}\mathbf{U}\bm{x_2}+\bm{b})  \\
    \mathrm{Biaffine}(\bm{x_1},\bm{x_2}) &=
    \mathrm{\sigma}(\bm{x_1}\mathbf{U}\bm{x_2}+\mathbf{W}[\bm{x_1};\bm{x_2}] +\bm{b})
\end{split}
\nonumber
\end{equation}
where $\mathbf{U}$ and $\mathbf{W}$ are the parameter matrix, $\bm{b}$ is the bias. $\mathrm{\sigma}$ is the sigmoid operation, guaranteeing that each governing score is between 0 and 1.

\begin{figure}[t]
\centering
\includegraphics[width=0.48\textwidth]{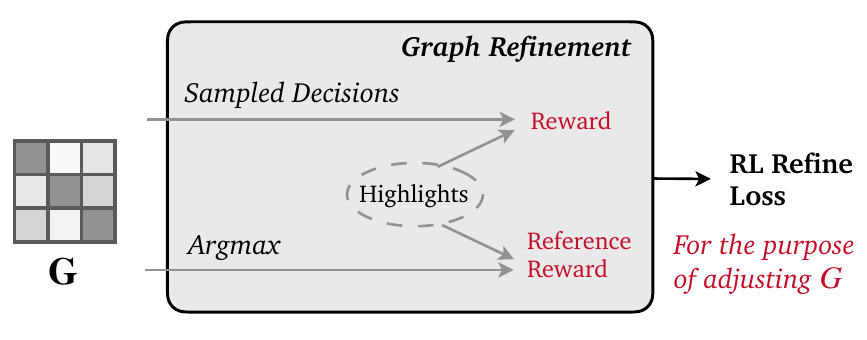}
\caption{The idea of the graph refinement module aims to adjust the governing scores in $\mathbf{G}$ with the help of highlights. Sampling decisions and greedily selecting by $\mathrm{argmax}$ are both consistent with the mind-map detector (\S\ref{subsec:mind_map_detector}). This builds a bridge between the learning of a graph (Phase \uppercase\expandafter{\romannumeral1}) and extracting a mind-map from the graph (Phase \uppercase\expandafter{\romannumeral2}).}
\label{Network_GR}
\end{figure}

\subsection{Graph Refinement}
\label{subsec:graph_refinement}
According to \cite{buzan2006}, a mind-map is organized as a tree structure with the central concept as its root node, and the major ideas are connected directly to the root. Other ideas branch out from the major ideas. Therefore, a clear mind-map tends to organize the main opinions of a document close to the root node. To achieve this goal, we leverage the human-written highlights to refine the graph $\mathbf{G}$ via reinforcement learning (RL) algorithm \cite{williams1992simple}, more specifically, self-critical RL \cite{rennie2017self}. The main idea is depicted in Figure \ref{Network_GR}. 

Concretely, the graph detector module can be considered as an \emph{agent} that follows a policy function to decide an action given a state. We regard an input document as the \emph{state} and the extracted graph $\mathbf{G}$ as the \emph{action} distribution. After we sample selected sentences over the graph, a delayed reward is calculated that indicates the similarity between selected sentences and highlights. Maximizing the expected reward helps to refine the governing scores in the graph $\mathbf{G}$. Next, we will introduce the graph refinement module in detail.

\noindent
{\bf Policy} \; The policy is described as below.
\begin{equation}
    \text{Sampled Decisions} \sim \pi_{\Theta}(\mathbf{G}|D)
    \label{eq:sampled_decisions}
\end{equation}
where $\Theta$ is the parameters of the graph detector, $D$ is the document and $\mathbf{G}$ is the extracted graph. We sample sentences over the graph as follows.

\noindent
{\bf Sampled Decisions} \; The main reason why RL can improve the reward is that it accords with the \emph{trial-and-error} process, which samples and update the parameters accordingly. To bridge with the strategy in the second phase (\S\ref{subsec:mind_map_detector}), i.e. detecting mind-map from a graph, we sample the upper nodes given the graph in the same way. At first, we sample a sentence as the root node of the mind-map.
\begin{equation}
    \bm{g_0} = \mathrm{softmax}(\mathrm{rowsum}(\mathbf{G}))
\end{equation}
where $\mathrm{rowsum}$ is row-wise summation. Its result is the salience score that a sentence governs all other sentences. A larger salience score indicates that a sentence is more likely to represent the key ideas of the document. We sample a root node based on multinomial distribution $\bm{g_0}$. Next, we remove the sampled root from the graph and cluster the remaining nodes into two sets, obtaining two subgraphs, i.e., $\mathbf{G}_1$ and $\mathbf{G}_2$. Similar with the root node, its two child nodes are also sampled based on the distributions $\bm{g_1}=\mathrm{softmax}(\mathrm{rowsum}(\mathbf{G}_1))$ and $\bm{g_2}=\mathrm{softmax}(\mathrm{rowsum}(\mathbf{G}_2))$, respectively.

The reason why we sample three sentences is that the average number of sentences in highlights is around 3.55. We also found that sampling more nodes does not improve performance. A possible reason is that more upper nodes introduce noise when comparing with highlights.

\noindent
{\bf Reward} \; The definition of reward is crucial for RL as it determines the optimizing objective. To ensure that the upper nodes of the mind-map represent the central concept and major ideas of the document, we treat the manual highlights as a reference. The ROUGE score \cite{lin-2004-rouge} between the sampled decisions and the highlights $A$ is used to define the reward. Multiple variants of the ROUGE score are proposed \cite{lin-2004-rouge}. Among them, ROUGE-1 ($\mathrm{R}\text{-}\mathrm{1}$), ROUGE-2 ($\mathrm{R}\text{-}\mathrm{2}$), ROUGE-L ($\mathrm{R}\text{-}\mathrm{L}$) are the most commonly utilized. We employ the average of ROUGE variants to define a reward function.
\begin{equation}
\small
    \mathrm{Sim}(X, A)=\frac{\mathrm{R}\text{-}\mathrm{1}(X,A)+\mathrm{R}\text{-}\mathrm{2}(X,A)+\mathrm{R}\text{-}\mathrm{L}(X,A)}{3}
    \label{eq:similarity}
\end{equation}

Assume the sampled sentences are $D_s$ and $D_s\subseteq{D}$, the reward is computed as follows.
\begin{equation}
    r = \mathrm{Sim}(D_s, A)
\end{equation}

\noindent
{\bf RL Refine Loss} \; According to \cite{sutton2000policy}, RL training objective is to maximize the expected reward. Therefore, we define the RL loss for graph refinement is to minimize the negative reward (see Eq. (\ref{eq:refine_loss})). More concretely, assume the sampled sentences $D_s=\{a_0,a_1,a_2\}$, where $a_0$ is the root, $a_1$ and $a_2$ are independent child nodes. Based on the conditional independence \cite{dawid1979conditional}, we have
\begin{equation}
\nonumber
    p(D_s)=p(a_0a_1a_2)=p(a_0)p(a_1|a_0)p(a_2|a_0)
\end{equation}

Therefore, $p(D_s)=g_0g_1g_2$, where $g_i$ is the probability of the sampled sentence in $\bm{g}_i$. When we only sample one sentence as the root node, $p(D_s)=g_0$.
\begin{equation}
    \mathcal{L}_r=-r\cdot{p(D_s)}=-r\prod_i{g_i}
    \label{eq:refine_loss}
\end{equation}

To reduce the variance caused by sampling, we associate the reward with a reference reward or baseline \cite{rennie2017self} and define it as $b=\mathrm{Sim}(D_b,A)$. $D_b$ is the sentences chosen greedily by the $\mathrm{argmax}$ operation on the multinomial distributions. With the likelihood ratio trick, the optimizing objective can be formulated as.
\begin{equation}
    \mathcal{L}_r = -(r-b)\sum_i\mathrm{log}(g_i)
    \label{eq:loss_rl}
\end{equation}

\begin{algorithm}[t]
\caption{Graph Detector Training Process}
\hspace*{0.02in} {\bf Input:} Training data $\mathcal{B}$: $\{B^1,B^2,...,B^K\}$  
\begin{algorithmic}[1]
\State Randomly initialize $\Theta$ 
\Repeat
    \For{$B^k\in\mathcal{B}$}
        \State Calculate graphs by graph detector
        \State Initialize temp batch loss $\hat{\mathcal{L}}\leftarrow{0}$
        \For{each document in $B^k$}
            \State Compute $\mathcal{L}_g$ by Eq. (\ref{eq:loss_mse})
            \State Compute $\mathcal{L}_r$ by Eq. (\ref{eq:loss_rl})
            \State Compute joint loss $\mathcal{L}$ by Eq. (\ref{eq:loss_combine})
            \State Update temp batch loss $\hat{\mathcal{L}}\leftarrow\hat{\mathcal{L}}+\mathcal{L}$
        \EndFor
        \State Optimize $\Theta$ by $\mathcal{\hat{L}}/|B^k|$
    \EndFor
\Until \emph{performance on the validation set does not improve in 3 epochs}
\end{algorithmic}
\label{algorithm_train}
\end{algorithm}


\subsection{Training}
We train the graph detector module by a combination of two optimizing objectives, i.e., fitting the pseudo graphs annotated by DistilBert and refining the generated graphs. 

Since it costs too much to manually annotating the relation labels in the graph, we automatically annotate a pseudo graph $\mathbf{Y}$ by DistilBert \cite{sanh2019distilbert}. In advance, DistilBert is fine-tuned by sentence pairs constructed from news articles. In this way, our method can take advantage of the prior knowledge from the pre-trained model, but also the local semantic association of sentence pairs. The fine-tuning details of DistilBert will be introduced in \S\ref{subsec:experimental_settings_distilbert}. The proposed model fits the pseudo graph by a mean square error (MSE) loss.
\begin{equation}
    \mathcal{L}_g = \frac{1}{N^2}\sum_{i}\sum_{j}(\mathbf{G}_{i,j}-\mathbf{Y}_{i,j})^2
    \label{eq:loss_mse}
\end{equation}
where $\mathbf{Y}$ is the pseudo graph.

Then we combine the MSE loss and graph refinement as an overall training objective to optimize the parameters $\Theta$. The entire training process of the proposed model is described in Algorithm \ref{algorithm_train}.
\begin{equation}
    \mathcal{L} = \mathcal{L}_g + \lambda{\mathcal{L}_r}
    \label{eq:loss_combine}
\end{equation}
where $\lambda$ balances the effect of graph refinement.

\subsection{Mind-Map Detector} 
\label{subsec:mind_map_detector}
In this section, we introduce how to generate a mind-map from a graph, i.e. $\mathbf{G}\rightarrow{M}$. The graph $\mathbf{G}$ covers all the sentences in the document, which might be redundant. To highlight the major ideas, we convert the graph into a mind-map through the strategy proposed by \cite{ijcai2019-729} to prune the extra edges. The algorithm works recursively to determine the governing relationship of sentences. First, it chooses a governor by picking the highest row-wise aggregation scores in the graph. Then except for the governor, it clusters the remaining nodes into two sub-groups with $k$-means algorithm. The sub-groups are processed recursively to extract the final mind-map. We enclose the full algorithm in the Appendix.

We extract two types of mind-map, i.e. salient-sentence-based mind-map (SSM) and key-snippet-based mind-map (KSM). Given the graph $\mathbf{G}$ of a document, we first prune it into SSM, and then extract the key phrases in each sentence \cite{rose2010automatic} to obtain the KSM. Therefore, SSM and KSM have the same structure. The only difference is that a node in SSM is a sentence, while a node in KSM is the key phrases. In the case of KSM, if no key phrase is found, the whole sentence is kept in the mind-map.

\section{Experiments}

\begin{table*}[t!]
\small
\begin{center}
\setlength{\tabcolsep}{2.2mm}{
\begin{tabular} {ll|cccc|cccc}
\toprule
     \multicolumn{2}{c}{\multirow{2}{*}{Models}} & \multicolumn{4}{|c}{SSM} & \multicolumn{4}{|c}{KSM} \\
    & & R-1 & R-2 & R-L & Avg & R-1 & R-2 & R-L & Avg \\
    \midrule
    \multirow{4}{*}{\shortstack{Compared \\ Methods}} & Random & 32.71 & 23.51 & 30.08 & 28.77 & 29.73 & 26.50 & 29.67 & 28.63 \\
    & LexRank & 34.53 & 25.04 & 31.79 & 30.45 & 31.04 & 27.75 & 31.00 & 29.93 \\
    & MRDMF & 38.19 & 29.51 & 35.72 & 34.47 & 33.18 & 30.26 & 33.08 & 32.18 \\
    & DistilBert & 42.15 & 33.34 & 39.66 & 38.38 & 40.00 & 36.92 & 39.92 & 38.95 \\
    \midrule
    \multirow{4}{*}{\shortstack{Model \\ Variants}} & EMGN(root) & 46.04 & 38.05 & 43.73 & 42.61 & 43.28 & {\bf 40.69} & 43.23 & 42.40 \\
    & EMGN(root)+greedy & 43.27 &35.11 & 40.93 & 39.77 & 40.30 & 37.62 & 40.24 & 39.39   \\
    & EMGN-GR & 45.06$^\dag$ & 37.08$^\dag$ & 42.75$^\dag$ & 41.63$^\dag$ & 41.62$^\dag$ & 39.07$^\dag$ & 41.57$^\dag$ & 40.75$^\dag$ \\
    & EMGN(biaffine) & 45.73 & 37.62 & 43.35 & 42.23 & 42.90 & 40.15 & 42.84 & 41.96 \\
    \midrule
    Full Model & EMGN & {\bf 46.14}$^\ddag$ & {\bf 38.21}$^\ddag$ & {\bf 43.84}$^\ddag$ & {\bf 42.73}$^\ddag$ & {\bf 43.33}$^\ddag$ & 40.67$^\ddag$ & {\bf 43.28}$^\ddag$ & {\bf 42.43}$^\ddag$ \\
    
\bottomrule
\end{tabular}}
\end{center}
\caption{\label{table-result} Evaluation results of the  salient-sentence-based mind-map (SSM) and key-snippet-based mind-map (KSM) in terms of R-1 (\%), R-2 (\%), R-L (\%) and the average score (\%). The marker $^\dag$ refers to $p$-value<0.01 when comparing with DistilBert. The marker $^\ddag$ refers to $p$-value<0.01 when comparing with EMGN-GR.}
\end{table*}


\subsection{Dataset}
We build an evaluation benchmark with 135 articles, which are randomly selected from CNN news articles \cite{hermann2015teaching,cheng2016neural}. The size of the benchmark is about 98,181 words. The average length of the news article is about 727 words. Two experts manually annotate the ground-truth mind-maps for these articles. If one of the experts disagrees with any content of the mind-map, they discuss to reach consensus. In the experiments, the benchmark is split into two datasets: a testing set $\mathcal{D}_{t}$ with 120 articles and a validation set $\mathcal{D}_{v}$ with 15 articles.

\subsection{Experimental Settings} 
\label{subsec:experimental_settings}
\subsubsection{Automatically Annotate Graphs for EMGN Training}
\label{subsec:experimental_settings_distilbert}
{\bf Sentence Pairs for Fine-tuning DistilBert} \; To save time in fine-tuning and subsequent annotation, we choose DistilBert as the \emph{``annotator''} to obtain the relationships of all sentence pairs in the graph. 

To construct the training pairs for fine-tuning DistilBert, we first randomly select 90k CNN news articles $\mathcal{D}_{news}$, which has no overlap with the benchmark. Each news consists of content and highlights. Because highlights summarize the major concepts of the news, they are regarded as the governors. To find the sentence pairs with governing relationships, we exploit TFIDF as the similarity metric. Concretely, a highlight governs each sentence in a paragraph when it is similar to one or some sentences in the paragraph. The negative samples are generated randomly.

In this way, we build a large-scale training corpus, which has 641,476 pairs from these news articles. Then we split all the pairs into 600k for training and 41,476 for testing.  

\noindent
{\bf Fine-tuning DistilBert} \; Using the training pairs, we fine-tune DistilBert for 3 epochs with a learning rate of 5e-5 and a training batch size of 32. The accuracy and F1 on the testing pairs are both more than 99.35\%. Thus DistilBert can annotate pseudo graphs with high confidence.


\noindent
{\bf Annotate Pseudo Graphs} \; We select 44,450 articles from $\mathcal{D}_{news}$ by setting the max length of sentence in the article as 50 and max number of sentences as 50\footnote{According to our statistical analysis on the CNN dataset, the average length of articles is 33.87 sentences and the average length of sentences is 21.25 words.}. After annotating these articles by DistilBert, they are exploited to train our mind-map generation model EMGN. 


\subsubsection{Mind-Map Evaluation}
We evaluate the generated mind-map by \emph{comparing the tree similarity with the human-annotated mind-map} \cite{ijcai2019-729}. We first remove the weak edges from a generated mind-map to ensure that it has the same number of edges as the annotation. The similarity between two edges is computed as below. We utilize $\mathrm{Sim}$ as Eq. (\ref{eq:similarity}).
\begin{equation} 
\nonumber
    {f}(s_i\rightarrow{s_j},s_a\rightarrow{s_b})=\frac{\mathrm{Sim}(s_i,s_a)+\mathrm{Sim}(s_j,s_b)}{2}
\end{equation}

Then for each edge in the annotation, the strategy finds the most similar edge in the generated mind-map. The final score is the average similarity score of all greedily selected pairs. 


\subsubsection{Implementation Details}
We initialize the word embeddings with 50-dimension GloVE \cite{pennington2014glove} and fine-tune during training. All other parameters are initialized by sampling from a normal distribution of $\mathcal{N}(0,0.02)$. The hidden size of BiLSTM is set to be 25$\times$2. The models are optimized by Adam \cite{kingma2014adam} with a learning rate of 1e-4. The batch size is 64. And $\lambda$ is set to 0.01. We employ an early stop strategy during training if the evaluation score on the validation set $\mathcal{D}_v$ does not improve in 3 epochs, and the best model is selected for evaluating testing set $\mathcal{D}_t$. For all baselines and our model, the reported results are the average score of 5 runs. 

The full results are presented in the Appendix.

\subsection{Experimental Methods}
\textbf{Compared Methods} \; We validate the effectiveness of the proposed method by comparing with the following baselines.
\begin{itemize}
    \vspace{-3pt}
    \item \textbf{Random}: We randomly sample a graph $\mathbf{G}$ for an input document. Each governing score $\mathbf{G}_{i,j}$ ranges from zero to one.
    \vspace{-3pt}
    \item \textbf{LexRank}: It computes the governing score of sentence pair by the cosine similarity of their TFIDF vectors. It follows the well-known LexRank algorithm \cite{erkan2004lexrank}, which is an extension of PageRank in the document summarization domain.
    \vspace{-3pt}
    \item \textbf{MRDMF} \cite{ijcai2019-729}: This is the state-of-the-art semantic mind-map generation work. It presents a multi-perspective recurrent detector to extract the governing relationship and then prunes the extra edges. 
    \vspace{-3pt}
    \item \textbf{DistilBert} \cite{sanh2019distilbert}: It is a lighter version of BERT \cite{devlin2019bert}. It provides the pseudo graphs for our method training. 
\end{itemize}

\noindent
\textbf{Method Variants} \; The proposed full model is the efficient mind-map generation network (EMGN), with bilinear operation in the sequence-to-graph module. We explore the impact of individual modules by comparing with its variants. Minus (-) means removing the module from the full model.
\begin{itemize}
    \vspace{-3pt}
    \item \textbf{EMGN(root)}: It only samples root node for refining the graph. 
    \vspace{-3pt}
    \item \textbf{EMGN(root)+greedy}: It chooses root node by greedily selecting the sentence with maximum similarity with highlights.
    \vspace{-3pt}
    \item \textbf{EMGN-GR}: It removes the graph refinement (GR) module from EMGN, which lefts the graph detector module for the purpose of sequence-to-graph.
    \vspace{-3pt}
    \item \textbf{EMGN(biaffine)}: It computes the graph by biaffine operation in EMGN.
\end{itemize}

\subsection{Experimental Results}
\label{subsec:experimental_results}


{\bf Overall Results} \; The experimental results for two types of mind-maps are displayed in Table \ref{table-result}. Firstly, we find that our method EMGN significantly outperforms MRDMF by 8.26\% on SSM and 10.25\% on KSM in terms of the average score. This indicates the effectiveness of EMGN. Then comparing DistilBert and EMGN, we can see that EMGN achieves significant improvements. This shows that the proposed method successfully exploits the pseudo labels and further improves the performances. Finally, we observe that DistilBert consistently outperforms MRDMF. Thus DistilBert is more effective in matching sequences than the multi-hop attention of MRDMF. Annotating pseudo graphs by DistilBert has higher confidence, which contributes to the subsequent learning of the proposed approach. 





\begin{table}[t!]
\small
\begin{center}
\setlength{\tabcolsep}{2.5mm}{
\begin{tabular} {l|cc|cc}
\toprule
    \multirow{2}{*}{Models} & \multicolumn{2}{c|}{SSM Avg} & \multicolumn{2}{c}{KSM Avg} \\
    & $\leq$ 25 & $>$ 25 & $\leq$ 25 & $>$ 25 \\
    \midrule
    Random & 31.94 & 26.63 & 32.53 & 26.22 \\
    LexRank & 33.54 & 28.07 & 33.99 & 27.28 \\
    MRDMF  & 39.83  & 30.58 & 36.76 & 28.69 \\
    DistilBert & 44.36 & 34.34 & 44.27 & 34.89 \\
    \midrule
    EMGN(root) & 49.91 & 37.32 & 49.39 & 37.24 \\
    EMGN-GR    & 49.22 & 36.22 & 46.81 & 36.14 \\
    EMGN(biaffine) & 49.54 & 37.00 & 48.36 & 37.13  \\
    EMGN  & {\bf 50.23} & {\bf 37.38} & {\bf 49.48} & {\bf 37.44} \\
\bottomrule
\end{tabular}}
\end{center}
\caption{\label{table-result-split} Evaluation results by splitting the testing set with the number of sentences in a document. Among all 120 files in $\mathcal{D}_t$, there are 50 files with the number of sentences $\leq$ 25, and 70 files with the number $>$ 25.}
\end{table}

\noindent
{\bf Compare with Model Variants} \; We further investigate the impact of individual components (see the second part of Table \ref{table-result}). We observe that EMGN-GR significantly outperforms DistilBert. By leveraging the sequential information of the document and early stop strategy, EMGN-GR can prevent overfitting the pseudo graphs and extract better mind-maps than DistilBert. By comparing EMGN-GR and EMGN, we found that EMGN significantly outperforms EMGN-GR. This verifies that the graph refinement module can successfully refine the graph to obtain a meaningful mind-map. 


Then, by comparing EMGN(root) and EMGN(root)+greedy, we see that EMGN(root) gains many improvements. A possible reason is that EMGN(root)+greedy only greedily enlarging the governing scores in the graph for a specific sentence, i.e. the one which has the maximum ROUGH similarity with highlights. This might ignore the exploration on other nodes. EMGN(root) performs better by sampling more sentences and compares them relatively. Finally, we found that EMGN performs slightly better than EMGN(root). This shows that refining more upper nodes achieves better performance than only refining the root node.


\noindent
\textbf{Effects of the Document Length} \; 
Table \ref{table-result-split} displays the evaluation results by splitting the testing set $\mathcal{D}_t$ with the document length, i.e. the number of sentences in a document. We found that EMGN consistently achieves the best performances. By comparing the results on two subsets, we can see that the results are highly related to the number of sentences in the article. It is still very challenging to extract meaningful mind-maps for the longer articles.

\begin{table}[t!]
\small
\begin{center}
\setlength{\tabcolsep}{2.5mm}{
\begin{tabular} {l|cc}
\toprule
    Models & $\mathcal{D}_t$ & $\mathcal{D}_v$  \\
    \midrule
    LexRank & 349.21 & 95.24 \\
    MRDMF & 467.07 & 62.49 \\
    DistilBert & 1219.51 & 201.04  \\
    \midrule
    EMGN-GR & 0.13 & 0.02 \\
    EMGN(root) & 3.18 & 0.44 \\
    EMGN(root) w/o sample & 0.14 & 0.02 \\
    EMGN & 10.27 & 1.42 \\
    EMGN w/o sample & 0.15 & 0.02 \\
\bottomrule
\end{tabular}}
\end{center}
\caption{\label{table-result-time} Total inference time (second) in Phase \uppercase\expandafter{\romannumeral1} of different methods. ``w/o sample'' indicates that removing the sampling in graph refinement module since we do not need to sample during evaluation.}
\end{table}

\subsection{Further Analysis}
{\bf Inference Time} \; To validate the efficiency of the proposed method, we compare the inference time of the testing set $\mathcal{D}_t$ and validation set $\mathcal{D}_v$ (see Table \ref{table-result-time}). Since all methods share Phase \uppercase\expandafter{\romannumeral2}, we only report the inference time of Phase \uppercase\expandafter{\romannumeral1} in Table \ref{table-result-time} to show the merits of the proposed method. The total inference time of Phase \uppercase\expandafter{\romannumeral2} is around 23.54 seconds in $D_t$ and 2.94 seconds in $D_v$.

We set the batch size of MRDMF and DistilBert as 256 (256 sentence pairs in a batch). The batch size of EMGN related methods is 32 (32 documents in a batch). We observe that the inference time of the existing methods, e.g. MRDMF and DistilBert, are more than 3,000 times compared with our method. As depicted in Figure \ref{example_method}, the main reason is that we significantly reduce the computational complexity of a relation graph from the sentence pair level to the document level. 



\begin{figure}[t]
\centering
\includegraphics[width=0.48\textwidth]{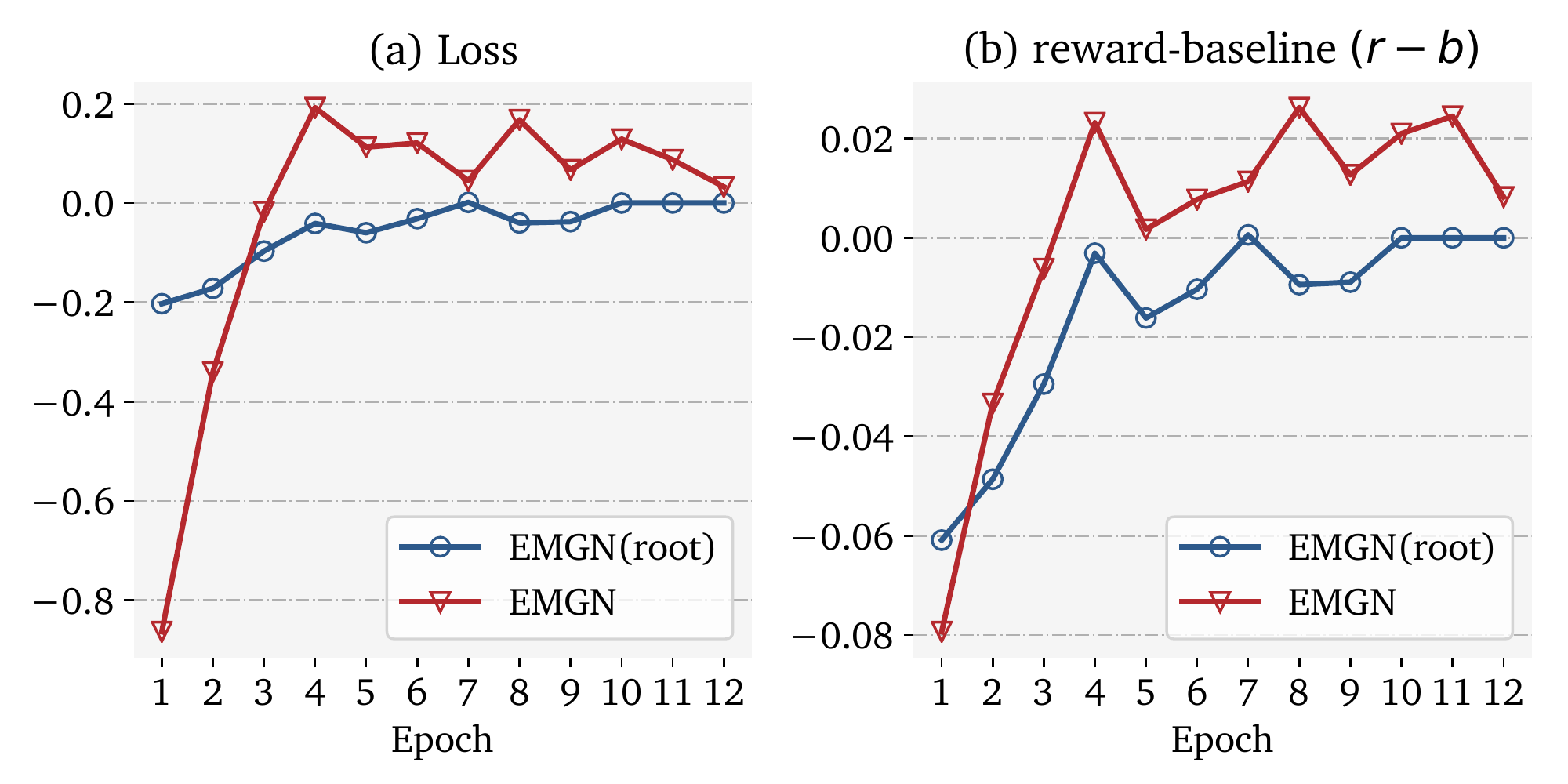}
\caption{The loss and reward curves in the graph refinement module.}
\label{figure-loss-reward}
\end{figure}

\noindent
{\bf Loss and Reward in Graph Refinement} \; The graph refinement aims to optimize the upper nodes of the mind-map for better revealing the major ideas of the document. We achieve this goal by optimizing the reward of sampled decisions. In Figure \ref{figure-loss-reward}, we plot the average loss $\mathcal{L}_r$ and average reward (Eq. \ref{eq:loss_rl}) in each epoch of the training process. In Figure \ref{figure-loss-reward}(a), it can be seen that the loss $\mathcal{L}_r$ gradually converges as training in both EMGN(root) and EMGN. Also in Figure \ref{figure-loss-reward}(b), the reward are gradually increasing as the training epochs and finally reach a relatively stable value. The training curves further prove that the proposed graph refinement module helps improve the similarity between the upper nodes and human-written highlights.

\begin{figure}[t]
\centering
\includegraphics[width=0.48\textwidth]{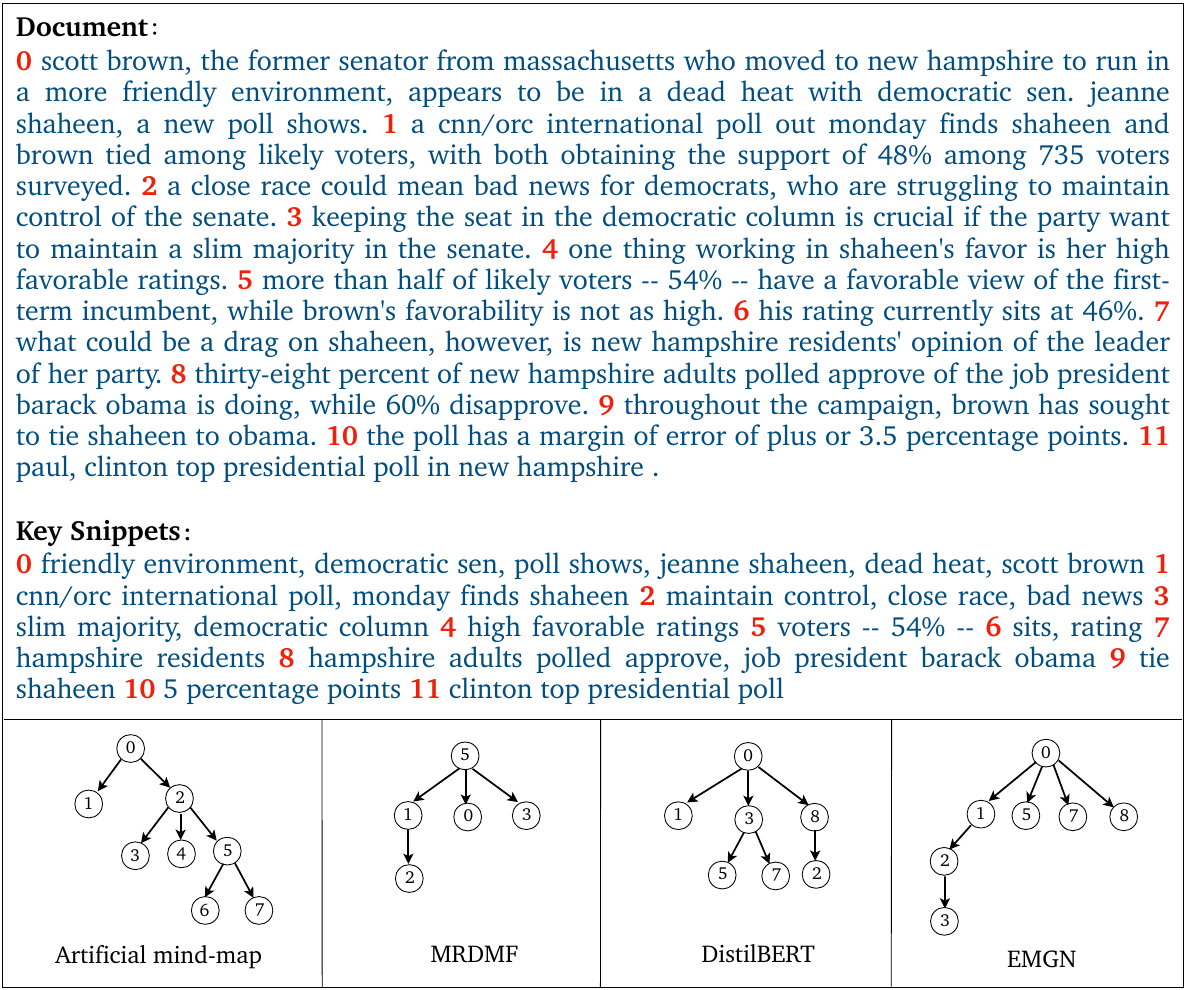}
\caption{Case study for SSM and KSM.}
\label{figure-case}
\end{figure}

\noindent
{\bf Case Study} \; In Figure \ref{figure-case}, we depict the generated mind-maps by varied methods for a CNN news\footnote{Article is available at \emph{\url{https://www.cnn.com/2014/09/15/politics/new-hampshire-senate-poll/index.html}}}. Comparing with the artificial mind-map, MRDMF chooses an inaccurate sentence as the root node. DistilBert and EMGN both generate the mind-map which represents the major ideas of the document. However, some relations between nodes in DistilBert are meaningless, such as the governing relation from sentence 8 to sentence 2. EMGN generates a mind-map that captures the central concept and grasps the directed relations among sentences. This is because our method considers the sequential information of an article and understands it as a whole. The case study further verifies that our method effectively generates a mind-map to reveal the underlying semantic structure for a document. 



\section{Related Works}
A mind-map is a hierarchical diagram to reveal the semantic structure for a document. The salient-sentence-based mind-map (SSM) is similar with extractive summarization \cite{zhong-etal-2020-extractive}, which aims to choose some key sentences from the document to describe the main ideas. Similar but completely different, mind-map generation can reveal not only the main ideas, but also the key semantic logic structure of the document. 

One previous work LexRank \cite{erkan2004lexrank} computes an adjacency matrix of the graph representation of sentences based on intra-sentence cosine similarity. However, the lexical similarity of some sentence pairs with semantic relation may be zero. Generating a mind-map from this graph representation tends to be less-meaningful, which is also indicated in the experiments (see \S\ref{subsec:experimental_results}). In addition, a few extractive summarization works employ graph techniques. For instance, a bipartite graph for sentence and entity nodes \cite{Parveen2015} or a weighted graph with topic nodes \cite{parveen-etal-2015-topical} are proposed to improve ranking the sentences in a document. Recently, \newcite{wang-etal-2020-heterogeneous} propose to build a heterogeneous graph to learn the correlation between word and sentence nodes, which helps to select better-summarizing sentences. Though these works involve learning the graph knowledge, such graphs are hard to derive a mind-map that focuses on the governing relationships between sentences.

Another related direction is the policy-based reinforcement learning \cite{williams1992simple,sutton2000policy}. Previous methods \cite{xiong2017deeppath,Xiao_Tan_Fan_Xu_Zhu_2020} usually affect the training of the main task by a policy network with separate parameters. Different from them, we directly regard the main network as the policy network and its output graph as the action distribution. Then the main network is optimized simultaneously when maximizing the expected reward. 

\section{Conclusion}
We propose an efficient mind-map generation network that converts a document into a graph via sequence-to-graph. To ensure a meaningful mind-map, we design a graph refinement module to adjust the graph by leveraging the highlights in a reinforcement learning manner. Extensive experimental results demonstrate that the proposed approach is more effective and efficient than the existing methods. The inference time is reduced by thousands of times compared with the existing approaches. The case studies further verify that the generated mind-map can reveal the underlying semantic structures of a document.

\section*{Acknowledgements}
We sincerely thank all the anonymous reviewers for providing valuable feedback. This work is supported by the National Science and Technology Major Project, China (Grant No. 2018YFB0204304).




\bibliography{custom}
\bibliographystyle{acl_bib}

\clearpage
\begin{appendix}

\begin{table*}[t]
\small
\begin{center}
\setlength{\tabcolsep}{4mm}{
\begin{tabular} {l|cccc}
\toprule
    \multirow{2}{*}{Models} & \multicolumn{4}{c}{SSM} \\
    & R-1 (\%) & R-2 (\%) & R-L (\%) & Avg (\%)  \\
    \midrule
    Random & 32.71 $\pm$ 0.44 & 23.51 $\pm$ 0.40 & 30.08 $\pm$ 0.40 & 28.77 $\pm$ 0.41 \\
    LexRank & 34.53 $\pm$ 0.06 & 25.04 $\pm$ 0.07 & 31.79 $\pm$ 0.06 & 30.45 $\pm$ 0.06 \\
    MRDMF & 38.19 $\pm$ 0.29 & 29.51 $\pm$ 0.35 & 35.72 $\pm$ 0.30 & 34.47 $\pm$ 0.31 \\
    DistilBert & 42.15 $\pm$ 0.19 & 33.34 $\pm$ 0.25 & 39.66 $\pm$ 0.21 & 38.38 $\pm$ 0.22 \\
    \midrule
    
    EMGN(root) & 46.04 $\pm$ 0.15 & 38.05 $\pm$ 0.20 & 43.73 $\pm$ 0.18 & 42.61 $\pm$ 0.17 \\
    EMGN(root)+greedy & 43.27 $\pm$ 0.87 &35.11 $\pm$ 1.01 & 40.93 $\pm$ 0.92 & 39.77 $\pm$ 0.93 \\
    EMGN-GR & 45.06 $\pm$ 0.64 & 37.08 $\pm$ 0.61 & 42.75 $\pm$ 0.63 & 41.63 $\pm$ 0.63 \\
    EMGN(biaffine) & 45.73 $\pm$ 0.70 & 37.62 $\pm$ 0.91 & 43.35 $\pm$ 0.79 & 42.23 $\pm$ 0.80 \\
    \midrule
    EMGN & {\bf 46.14 $\pm$ 0.15} & {\bf 38.21 $\pm$ 0.21} & {\bf 43.84 $\pm$ 0.17} & {\bf 42.73 $\pm$ 0.17} \\
    
\bottomrule
\end{tabular}}
\end{center}
\caption{\label{table-result-appendix} Full evaluation results of salient-sentence-based mind-map (SSM).}
\end{table*}

\begin{table*}[t]
\small
\begin{center}
\setlength{\tabcolsep}{4mm}{
\begin{tabular} {l|cccc}
\toprule
    \multirow{2}{*}{Models} & \multicolumn{4}{c}{KSM} \\
    & R-1 (\%) & R-2 (\%) & R-L (\%) & Avg (\%)  \\
    \midrule
    Random & 29.73 $\pm$ 0.51 & 26.50 $\pm$ 0.57 & 29.67 $\pm$ 0.51 & 28.63 $\pm$ 0.53 \\
    LexRank & 31.04 $\pm$ 0.33 & 27.75 $\pm$ 0.37 & 31.00 $\pm$ 0.34 & 29.93 $\pm$ 0.35 \\
    MRDMF & 33.18 $\pm$ 0.43 & 30.26 $\pm$ 0.38 & 33.08 $\pm$ 0.43 & 32.18 $\pm$ 0.41 \\
    DistilBert & 40.00 $\pm$ 0.47 & 36.92 $\pm$ 0.45 & 39.92 $\pm$ 0.48 & 38.95 $\pm$ 0.47 \\
    \midrule
    
    EMGN(root) & 43.28 $\pm$ 0.03 & {\bf 40.69 $\pm$ 0.17} & 43.23 $\pm$ 0.03 & 42.40 $\pm$ 0.07 \\
    EMGN(root)+greedy & 40.30 $\pm$ 0.61 & 37.62 $\pm$ 0.53 & 40.24 $\pm$ 0.61 & 39.39 $\pm$ 0.58 \\
    EMGN-GR & 41.62 $\pm$ 0.96 & 39.07 $\pm$ 0.84 & 41.57 $\pm$ 0.96 & 40.75 $\pm$ 0.92 \\
    EMGN(biaffine) & 42.90 $\pm$ 0.64 & 40.15 $\pm$ 0.77 & 42.84 $\pm$ 0.64 & 41.96 $\pm$ 0.68 \\
    \midrule
    EMGN & {\bf 43.33 $\pm$ 0.38} & 40.67 $\pm$ 0.39 & {\bf 43.28 $\pm$ 0.37} & {\bf 42.43 $\pm$ 0.38} \\
\bottomrule
\end{tabular}}
\end{center}
\caption{\label{table-result-keyword-appendix} Full evaluation results of key-snippet-based mind-map (KSM).}
\end{table*}

\begin{table*}[t!]
\small
\begin{center}
\setlength{\tabcolsep}{4mm}{
\begin{tabular} {l|cc|cc}
\toprule
    \multirow{2}{*}{Models} & \multicolumn{2}{c|}{SSM Avg} & \multicolumn{2}{c}{KSM Avg} \\
    & $\leq$ 25 & $>$ 25 & $\leq$ 25 & $>$ 25\\
    \midrule
    Random & 31.94 $\pm$ 1.84 & 26.63 $\pm$ 0.82 & 32.53 $\pm$ 1.87 & 26.22 $\pm$ 0.81 \\
    LexRank & 33.54 $\pm$ 0.25 & 28.07 $\pm$ 0.09 & 33.99 $\pm$ 0.27 & 27.28 $\pm$ 0.07 \\
    MRDMF  & 39.83 $\pm$ 0.18  & 30.58 $\pm$ 0.23 & 36.76 $\pm$ 0.24 & 28.69 $\pm$ 0.27  \\
    DistilBert & 44.36 $\pm$ 0.34 & 34.34 $\pm$ 0.14 & 44.27 $\pm$ 0.78 & 34.89 $\pm$ 0.39 \\
    \midrule
    
    EMGN(root) & 49.91 $\pm$ 0.39 & 37.32 $\pm$ 0.34 & 49.39 $\pm$ 0.49 & 37.24 $\pm$ 0.49 \\
    EMGN-GR    & 49.22 $\pm$ 0.48 & 36.22 $\pm$ 0.47 & 46.81 $\pm$ 1.05 & 36.14 $\pm$ 0.53 \\
    EMGN(biaffine) & 49.54 $\pm$ 0.68 & 37.00 $\pm$ 0.64 & 48.36 $\pm$ 0.56 & 37.13 $\pm$ 0.75  \\
    EMGN  & {\bf 50.23 $\pm$ 0.59} & {\bf 37.38 $\pm$ 0.24} & {\bf 49.48 $\pm$ 0.52} & {\bf 37.44 $\pm$ 0.61} \\
\bottomrule
\end{tabular}}
\end{center}
\caption{\label{table-result-split-appendix} Full evaluation results of splitting testing set with the number of sentences in a document.}
\end{table*}

\begin{figure}[t]
\centering
\includegraphics[width=0.48\textwidth]{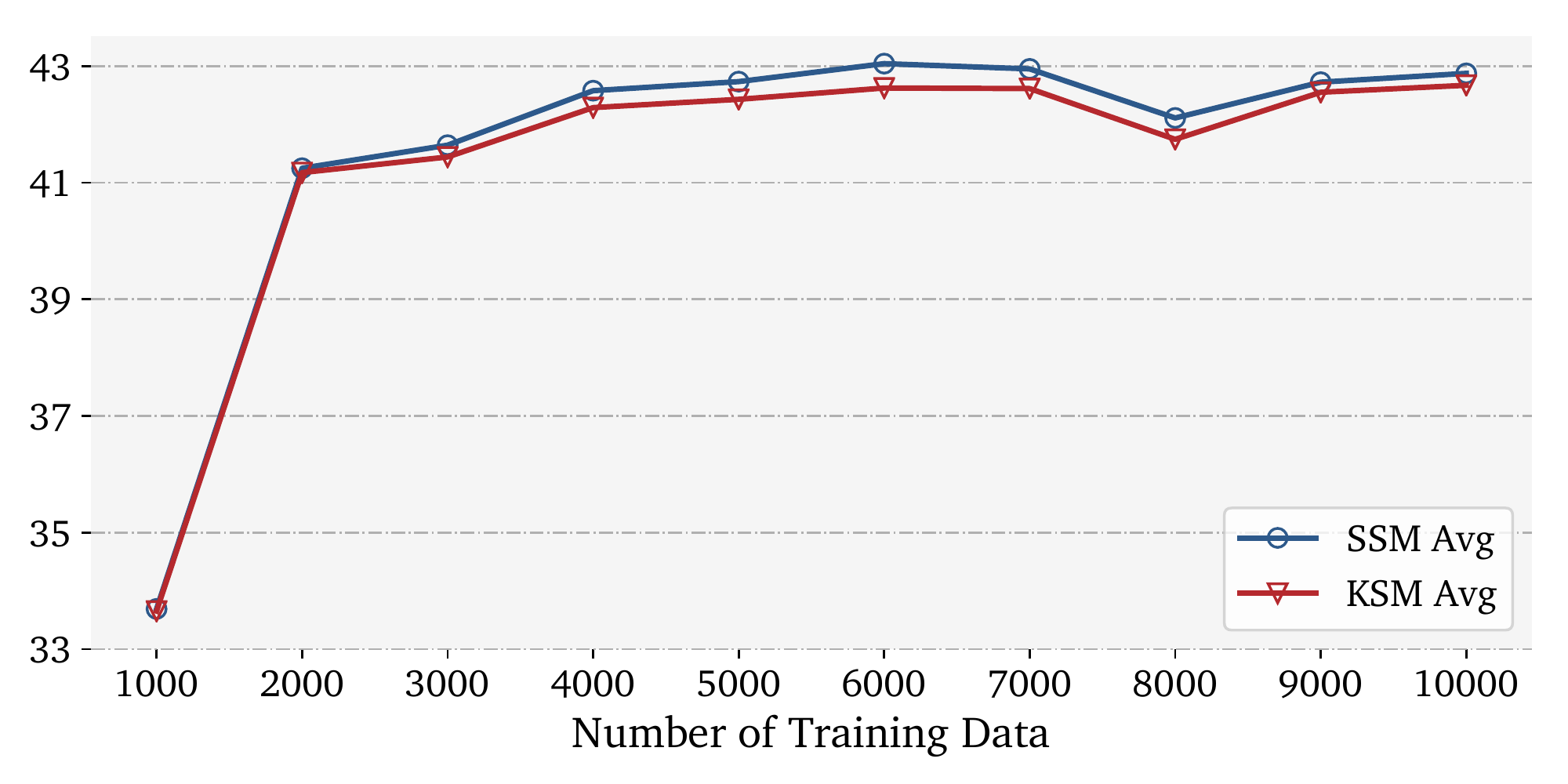}
\caption{Effects of training data scale for EMGN.}
\label{fig:training_data}
\end{figure}

\section{Software and Hardware}
We use Pytorch to implement all models (Python 3.5). The operating system is Red Hat Enterprise Linux 7.8. DistilBert is trained on Tesla K80. All other models are trained on GTX 980. 

We compare the inference time of all the models in the same software and hardware environments.



\section{Results Appendix}
\subsection{Full Results with Standard Deviations}
As shown in Table \ref{table-result-appendix}, Table \ref{table-result-keyword-appendix} and Table \ref{table-result-split-appendix}, we display the full experimental results, including the average score and the standard deviation of 5 runs. 

\begin{figure}[t]
\centering
\includegraphics[width=0.48\textwidth]{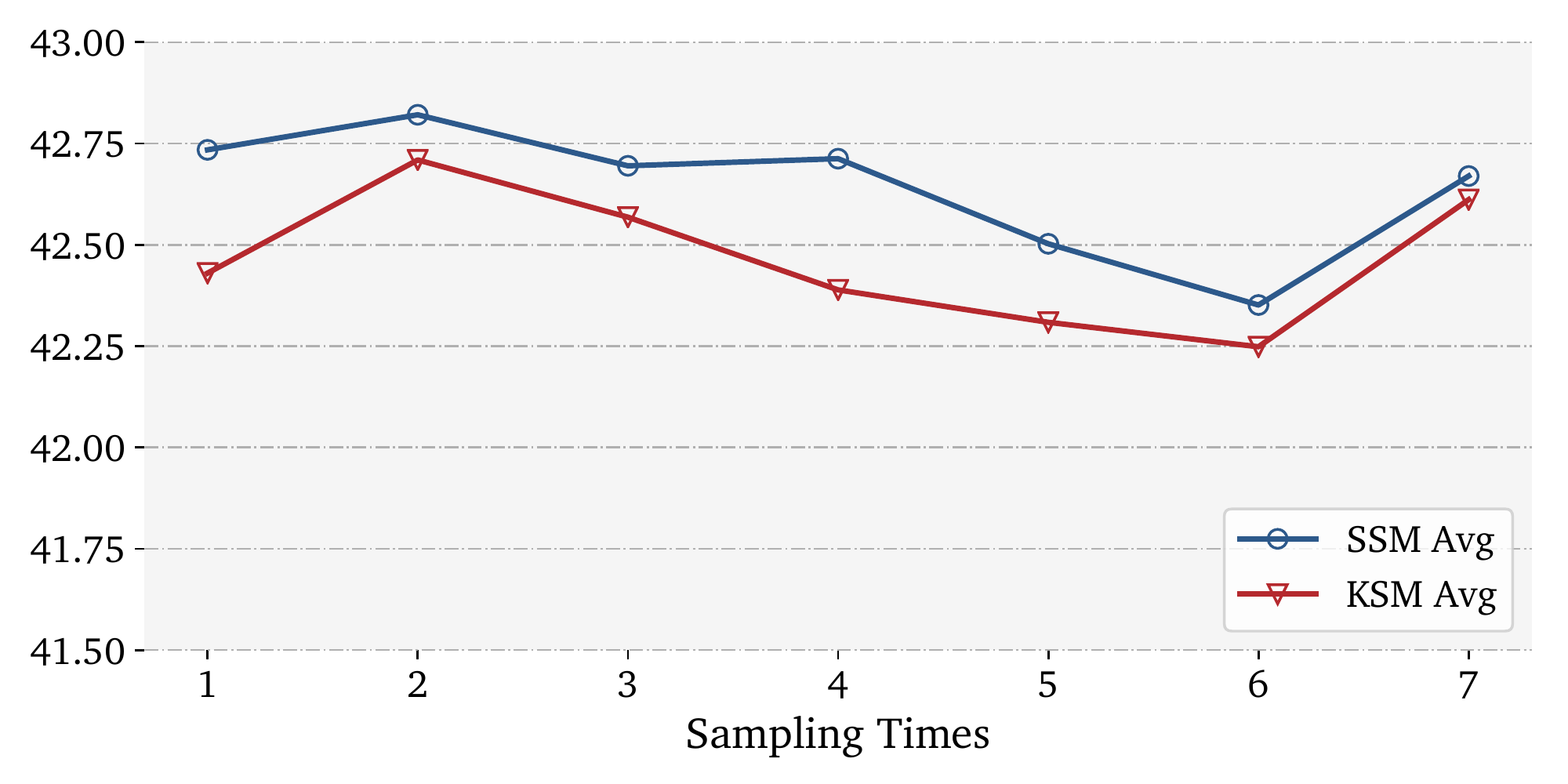}
\caption{Effects of sampling times for EMGN.}
\label{fig:sampling_times}
\end{figure}

\subsection{Effects of Training Data Scale}
We also investigate the impact of the training data size on the performance. We totally annotate pseudo graph labels for 44,450 documents by DistilBert. The performance curves of EMGN with different training scale are depicted in Figure \ref{fig:training_data}. It shows that training the proposed model EMGN does not require too many labeled documents. The performance scores are improved significantly by changing the data scale from 1000 to 2000. The results grow steadily as adding more training data. A possible explanation is that compared with the ground-truth graph, the pseudo graph labels by DistilBert are still less accurate and might have redundant patterns. 


Therefore, in the experiments section, all our models are trained with 5k labeled articles. 

\begin{algorithm}[t]
\caption{Salient-Sentence-based Mind-map Generator.}
\hspace*{0.02in} {\bf Require:} a document $D$ \\
\hspace*{0.02in} {\bf Ensure:} the nodes $\mathbf{C}_s$, and the edges $\mathbf{E}_s$
\begin{algorithmic}[1]
\Function{recursive}{$\mathbf{G}$,$\mathbf{C}$,$\mathbf{E}$,$root$} 
    \State $k$ $\leftarrow$ $length$($\mathbf{G}$)
    \State $\bm{g}$ $\leftarrow$ $\mathrm{rowsum}(\mathbf{G})$
    \State $governor \leftarrow s_i,s.t.,\bm{g}(i)=\mathrm{max}(\bm{g})$
    \If{$k>0$ and ($k=1$ or $\bm{g}(i)/k>$0.5)}
    \State $\mathbf{C}\leftarrow\mathbf{C}\cup\{governor\}$
    \State $\mathbf{E}\leftarrow\mathbf{E}\cup\{(root, governor)\}$
    \State $\mathbf{G}^{'}\leftarrow\mathbf{G}-\{\mathbf{G}_i\}$
    \State $root\leftarrow{governor}$
    \EndIf
    \If{$k\leq$1} \textbf{return} \EndIf
    \State $\mathbf{G}_1$, $\mathbf{G}_2\leftarrow$ clustering($\mathbf{G}^{'}$,2)
    \State recursive($\mathbf{G}_1$,$\mathbf{C}$,$\mathbf{E}$,$root$)
    \State recursive($\mathbf{G}_2$,$\mathbf{C}$,$\mathbf{E}$,$root$)
\EndFunction
\Function{Generator}{$D$}
    \State Obtain graph $\mathbf{G}$ for document $D$
    \State $\mathbf{C}_s\leftarrow\varnothing$, $\mathbf{E}_s\leftarrow\varnothing$
    \State recursive($\mathbf{G}$, $\mathbf{C}_s$, $\mathbf{E}_s$, $Null$)
    \State \textbf{return} $\mathbf{C}_s$, $\mathbf{E}_s$
\EndFunction
\end{algorithmic}
\label{algorithm_mindmap_generator}
\end{algorithm}

\subsection{Effects of Sampling Times}
In the graph refinement module, we sample the upper nodes and improve their ROUGE similarity with the human-written highlights. Figure \ref{fig:sampling_times} shows the performance curves of different sampling times for each document. We found that more sampling times do not improve the performances of EMGN significantly, while requires much more time to train the model. Therefore, in the experiments section, we set the sampling times as one for both EMGN(root) and EMGN.

\section{Mind-Map Generator}
The algorithm for mind-map generator is enclosed in Algorithm \ref{algorithm_mindmap_generator}. After calculating the graph $\mathbf{G}$ for a document, the nodes and edges are built recursively. It is worth noting that the graph refinement module also follows this recursive way.

We also enclose the algorithm for mind-map evaluation in Algorithm \ref{algorithm_mindmap_evaluation}. 

\begin{algorithm}[t]
\caption{Evaluation for Mind-map.}
\hspace*{0.02in} {\bf Require:} the edges of the manual mind-map $\mathbf{E}_1$, the edges of the generated mind-map $\mathbf{E}_2$ \\
\hspace*{0.02in} {\bf Ensure:} the edge similarity score $f$, and the most similar pair $ms$
\begin{algorithmic}[1]
\Function{simFunction}{$\mathbf{E}_1$, $\mathbf{E}_2$} 
    \State $r\leftarrow{0}$
    \State truncate($\mathbf{E}_2$),$s.t.|\mathbf{E}_2|=|\mathbf{E}_1|$
    \For{($s_i\rightarrow{s_j)}$ in $\mathbf{E}_1$}
        \State $ms\leftarrow{(None, None)}$
        \For{($s_a\rightarrow{s_b})$ in $\mathbf{E}_2$}
            \If{$f(s_i\rightarrow{s_j},ms[0]\rightarrow{ms[1]})$ \; \State <$f(s_i\rightarrow{s_j},s_a\rightarrow{s_b})$}
            \State $ms\leftarrow(s_a\rightarrow{s_b})$
            \EndIf
        \EndFor
        \State $r\leftarrow{f}(s_i\rightarrow{s_j},ms[0]\rightarrow{ms[1]})+r$
        \State $\mathbf{E}_2\leftarrow{\mathbf{E}_2}-ms$
    \EndFor
    \State \textbf{return} $r/|\mathbf{E}_1|$
\EndFunction
\end{algorithmic}
\label{algorithm_mindmap_evaluation}
\end{algorithm}

\end{appendix}
\end{document}